# ELM Solutions for Event-Based Systems

Jonathan Tapson [a,*], Gregory Cohen[*] and André van Schaik [a]

[a] *The MARCS Institute, University of Western Sydney, Penrith NSW, Australia 2751*

**Abstract**

Whilst most engineered systems use signals that are continuous in time, there is a domain of systems in which signals consist of events, or point processes. Events, like Dirac delta functions, have no meaningful time duration. Many important real-world systems are intrinsically event-based, including the mammalian brain, in which the primary packets of data are spike events, or action potentials. In this domain, signal processing requires responses to spatio-temporal patterns of events. We show that some straightforward modifications to the standard ELM topology produce networks that are able to perform spatio-temporal event processing online with a high degree of accuracy. The modifications involve the re-definition of hidden layer units as *synaptic kernels*, in which the input delta functions are transformed into continuous-valued signals using a variety of impulse-response functions. This permits the use of linear solution methods in the output layer, which can produce events or point processes as output, if modeled as a classifier with the two output classes being "event" or "no event". We illustrate the method in application to a spike-processing problem.

*Key words:* Extreme learning machine, event based processing, point processes, synaptic kernel, mixed domain processing, spatio-temporal pattern recognition, spatio-temporal receptive field.

* Corresponding author
  *Email address:* j.tapson@uws.edu.au (Jonathan Tapson).
  *Postal address:* University of Western Sydney, Locked Bag 1797, Penrith NSW 2751, Australia
  *Phone:* +61414373961

# 1 Introduction

The Extreme Learning Machine is well known for its performance and utility in standard regression and classification problems. A large part of its appeal is in the ease of network synthesis and weights solution [1]. This makes it an excellent candidate framework for some real-world problems in which network synthesis is hampered by a lack of fundamental knowledge of the system being modeled.

Most of the work in signal and network analysis is applied in the domain of signals that are intrinsically continuous in time, albeit that they may have been quantized in amplitude and time by a sampling process. Event-based systems, sometimes referred to as Lebesgue sampling systems [2], are significantly different in that events are Boolean in nature and have infinitesimally short time duration; they are usually modeled by Dirac delta functions, and statistically can be considered as point processes. As such, methods that have their origins in continuous-valued signal processing are seldom useful in processing events.

There are a number of good reasons for wanting to model event-based systems. One is that these systems are intrinsically economical in energy use, as they may be adjusted or tuned to produce no events in a steady state, but only to produce event signals when the state changes (these are sometimes referred to as "spike on delta" systems). An analysis by Astrom and Bernhardsson [2] indicates that event-based feedback control loops are more efficient, typically by a factor of 4.7 times, than sampled-data control loops, for the simple reason that no signals are sent, and no control action takes place, when the system requires no change.

A second reason for modeling event-based processes is that many real-world systems are event-based, and these include the human brain – possibly the most interesting unsolved modeling problem in science. Other event-based systems include radioactive decays, arrival of customers in a queue (and other economic processes in queueing theory), occurrence of earthquakes, and many more real-world processes where modeling may be important.

Most biological neurons signal other neurons by emitting action potentials, or spikes. It is axiomatic in neuroscience that spikes carry no information in their amplitudes or waveforms, and that it is their presence and timing that carry information [3-4]. The coding of information in spikes is characterized in terms of rate-, place-, and time-encoding. Rate-encoding implies that information is coded in the rate of production of spikes, and in many ways rate-coded spikes can be represented by a continuous rate variable, and then processed with conventional neural networks. Place-encoding refers to the concept that information is conveyed by which individual or subset of an ensemble of neurons spikes. Time-encoding infers that information is conveyed by the precise time of a spike event. Place and time encoding are generally combined into spatio-temporal coding, in which a particular pattern of spikes in time on a particular ensemble of neurons is considered to convey multi-dimensional information. There is considerable evidence for spatio-temporal encoding of information in many biological systems. Place-, time-, and spatio-temporally-encoded spikes are not generally suitable for processing by conventional neural networks, as each isolated spike contains or encodes information that is of only infinitesimal duration, and hence the signals do not have meaningful first derivatives. It is notable that energy efficiency favors the use of spatio-temporally coded information, as the use of a single spike to encode a scalar value is more efficient than the use of several spikes to define a rate which encodes a scalar value [5]. It remains a considerable challenge to computer engineers that the human brain operates with an energy consumption in the order of 25W [6], which is lower than most current PCs; there are many reasons for this energy efficiency, but the sparse use of spikes for information transmission is unarguably a major contributor.

This report is structured as follows: in Section 2 we review prior work on spatio-temporal event processing. In Section 3 we introduce our contribution to solving the problem, the Synaptic Kernel Inverse Method (SKIM), and in Sections 4 and 5 we provide detail on the synaptic kernels and the inverse solution methodology. In Section 6 we extend the method to mixed-domain systems which have both point processes (events) and continuous signals. In Section 7 we examine the method in the context of recurrently connected networks. Section 8 gives an example of the SKIM method applied to a mixed-domain problem, and illustrates many of the possibilities in its use. This is followed by some concluding remarks.

## 2  Previous Work on Spatio-Temporal Event Processing

Much of the prior work on event-based networks displays elements of the core ELM methodology, suggesting that this methodology lends itself to useful solutions of the problem. In particular, the use of randomly weighted links to connect an input layer to a higher-dimensional hidden layer, and the linear solution of weights between the hidden layer and the output, appear in a number of contexts.

There are a number of methods for spatio-temporal processing that are covered by the umbrella term of *reservoir computing*, including most notably the *liquid state machine* [7] and the *echo state network* [8]. These two methods were discovered in parallel in 2001, and have the same basic features: the input signals are fed to a reservoir interlayer of neurons that are recurrently connected, often in an all-to-all fashion, using random weights. The output is then "read out" by a linear output layer of neurons. There are some differences in the detail of the two approaches, particularly in the hidden layer neuron model, with liquid state machines being intended for use with spiking neurons and echo state networks using somewhat more conventional continuously valued signals. Notably, the output neurons may have feedback connections to the reservoir layer. Liquid state machines in particular have resonated with the computational neuroscience community and have become a popular model for postulates of brain connectivity, possibly more so in the conceptual than practical sense.

In a body of work that predates the liquid state machine, Maass and colleagues identified the potential for these types of network to have universal computational power, provided that conditions of *pointwise separability* and *fading memory* are met by the neuronal structures (called *filters* in Maass' work). Given that these same requirements underpin the conventional ELM, the work is of considerable relevance, and particularly so for this report. In particular, Maass notes that "A biologically quite interesting class of filters that satisfies the formal requirement of the pointwise separation is the class of filters defined by standard models for dynamic synapses," [7], [9]. The nonlinear synaptic kernels we introduce in Section 3 embody the depressive dynamics suggested by Maass and Sontag [9].

A further significant use of ELM-like principles in the neuroscientific context was by Eliasmith and Anderson, in the core algorithm of their Neural Engineering Framework (NEF) [10]. Subsequently, Eliasmith and colleagues have used the NEF platform to build extremely large cognitive networks, including most recently a network of 2.5 million neurons with apparent cognitive ability [11]. Basic NEF networks have spiking neurons as inputs, in the hidden layer, and as outputs. The weights between input and (large) hidden layers are randomized, and the output layers are linear and have weights which are solved by singular value decomposition. Memory of prior events is achieved by the use of recurrent connections in the hidden layer. A number of different spiking neuron models have been used in NEF, but the default is to use a leaky integrate-and-fire neuron (see e.g. Izhikevich [12] for a review of single-compartment spiking neuron models).

Despite its use of spiking neurons, the NEF intrinsically treats spikes as rate-encoded variables, so that for example the hidden-layer neurons are pre-characterized in terms of their input-output rate transfer function. It is nonetheless an extremely robust technique, and it shows some sensitivity to spike timing, although it is not likely that it is capable of useful results on signals where only a single spike per epoch per input channel might occur.

We have elsewhere characterized systems such as ELM, liquid state machines, echo state networks and the core NEF algorithm as *Linear Solutions of Higher Dimensional Interlayer* networks (LSHDI) [13]. Recognizing the convergent evolution of these methods allows us to make use of a depth of theoretical analysis of the common methodology. For example, the analysis of Rahimi and Recht [14] is applicable to all of these methods.

There are also feedforward-only methods of processing spatio-temporal patterns, based mostly of summing or subtracting exponentially-decaying event impulse responses [15-17]. Perhaps the best known is Gütig and Sompolinsky's Tempotron [17], which has inspired electronic circuit implementations such as the "Deltron" [18, 19].

In this paper we outline a modification to the standard ELM topology that enables processing of events in a way that intrinsically accommodates the sparse occurrences of events that are associated with spatio-temporal coding of information. The method is called the Synaptic Kernel Inverse Method (SKIM) and is outlined below.

## 3   The Synaptic Kernel Inverse Method

The SKIM topology is shown in Figure 1. The inputs are neurons which supply streams of events. These inputs are connected by random static weights to a hidden layer which consists of synapses. Each synapse implements a synaptic kernel function or filter, which represents the dynamic response of the synapse to an input event. The synaptic kernels are biologically inspired, being based on typical functions used to model the transfer functions in mammalian neurons. They are effectively filters which are defined in terms of their impulse response.

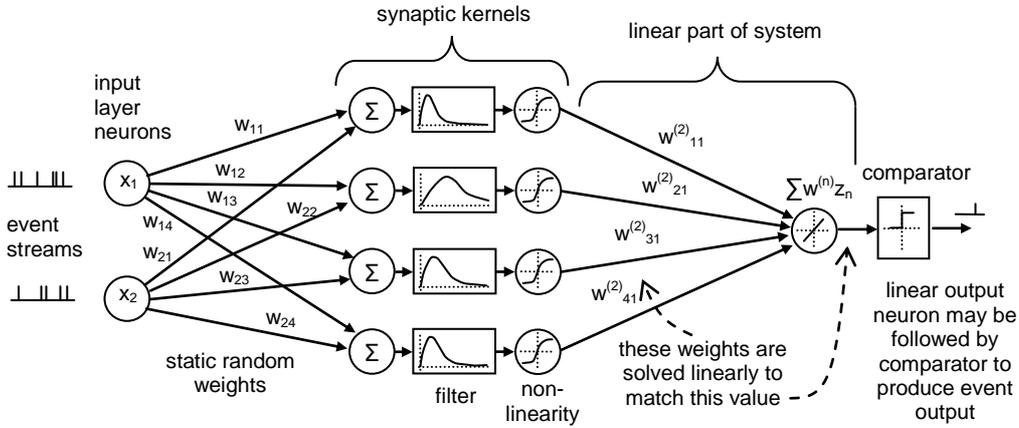

Figure 1. Topology of the Synaptic Kernel Inverse Method (SKIM), showing the basic ELM feedforward network adapted with synaptic kernels and event-based outputs.

The synaptic responses to input events are summed at the synapse and are also nonlinearly scaled, generally by means of a compressive nonlinearity such as the *tanh* function. The nonlinear summed synaptic responses constitute the outputs of the hidden layer.

The hidden layer outputs are summed by linear output neurons in the same way as conventional ELM. This provides the framework for linear solution of the output weights, by means of a pseudoinverse computation or similar optimization.

If the outputs of the networks are events, the output layer signals can be compared to a threshold, defining the event/no event boundary.

## 4  Synaptic Kernels

The synaptic kernels implement three functions:

- They convert the input events into continuous signals.
- They construct higher-dimensional features from the input event space.
- They provide short-term memory of past events.

Synaptic kernels have a lot in comment with wavelet functions, in that most candidates will perform reasonably, but the particular problem or model space will determine the optimal function. There are many possibilities, but the following characteristics are helpful:

- The impulse response of the function must have energy (non-zero amplitude) at times other than the input impulse time. For causal filters, this implies energy at times after the input event (post-spike times). This characteristic implements some short-term memory of the event.

- The peak amplitude of response should occur in a time scale of the same order as the typical interval between spikes in the target pattern, so that the signal energy is available to the output neurons in an appropriate time scale.
- The response should decay to zero in the long term (typically according to an exponential or power law). This contributes to the requirement for fading memory. For computational purposes it should be possible to truncate the response after some reasonable interval.
- The response should be nonlinear with respect to the summed impulse amplitudes, in order to improve the separability of the feature space over the space of linear combinations of input events. Compressive nonlinearities such as the logistic and *tanh* functions have the effect of creating a depressive synaptic adaptation, as analysed in terms of computational universality by Maass and Sontag [9].

A number of synaptic kernels are shown in Table 1. Generally, they consist of a filter function, such as a first-order low pass filter, a damped resonance, or an alpha function, followed by a compressive nonlinearity.

| Kernel Type | Expression for Impulse Response | Typical Function (Spikes at $t = 0$, $t = 100$) |
|---|---|---|
| Stable recurrent connection (leaky integration) with nonlinear leak | $g(t) = \frac{1}{1+(g(t-\tau))^2} \int_{t_0}^{t} \sum_{i=1}^{L} w_{ji}^{(1)} x_{i,t} \, dt$ | |
| Alpha function | $g(t) = \left[\sum_{i=1}^{L} w_{ji}^{(1)} x_{i,t}\right] \frac{t}{\tau} e^{-\frac{t}{\tau}}$ | |
| Damped resonant synapse | $g(t) = \left[\sum_{i=1}^{L} w_{ji}^{(1)} x_{i,t}\right] e^{-\frac{t}{\tau}} \sin(\omega t)$ | |
| Synaptic or dendritic delay with alpha function | for $t \geq \Delta t$: $g(t) = \left[\sum_{i=1}^{L} w_{ji}^{(1)} x_{i,t}\right] \frac{t-\Delta T}{\tau} e^{-\frac{t-\Delta T}{\tau}}$ $t < \Delta t: g(t) = 0$ | |
| Synaptic or dendritic delay with Gaussian function | $g(t) = \left[\sum_{i=1}^{L} w_{ji}^{(1)} x_{i,t}\right] \frac{1}{\sigma\sqrt{2\pi}} e^{-\frac{(t-\Delta T)^2}{2\sigma^2}}$ | |

Table 1. Examples of synaptic kernel or filter functions that can be used in the SKIM method. The graphs show the filter response for a single event at $t = 0$ (solid line) and a second event at $t = 100$ (dotted line); the events are shown in red. For the alpha and damped resonances, time constants were chosen for maximum energy at $t = 100$ steps, and for those with delays (the lowest two), at $t = 70$ steps. $\tau$ is the time constant for the various functions, $\Delta T$ is an explicit synaptic or dendritic delay, and $\omega$ the natural resonant frequency for a damped resonant synaptic function. Apart from the leaky integrator, in which the nonlinearity is inherent, the functions would need to be followed by a compressive nonlinearity such as a *tanh* or logistic function, as shown in Figure 1.

The SKIM network processes the input signals as follows: given a sequence of input events $\bar{x}_t \in \mathbb{R}^{L \times 1}$ on $L$ channels, where $t$ is a time or series index and $x_i \in \{0,1\}$ depending on whether there is an event at time $t$ or not, the synaptic filter functions $F(\cdot)$ operate on the events as follows:

$$S_j(t) = F_j\left(\sum_{i=1}^{L} w_{ji}^{(1)} x_{i,t}\right) \quad (1)$$

where the hidden layer weights $w_{ji}^{(1)}$ are initialized to static random values, usually uniformly distributed in some sensible range as in classical ELM. The superscript is a layer index. $S_j(t)$ is the output of the synaptic filter in kernel $j$ and is fed into a compressive nonlinearity (unless it has an intrinsic nonlinearity, as in a nonlinear leaky integrator). The $M$ outputs of the nonlinearly processed kernels are summed via the output weights at the $N$ output neurons, given by $y$, as:

$$y_n(t) = \sum_{j=1}^{M} w_{nj}^{(2)} \tanh\left( F_j \left( \sum_{i=1}^{L} w_{ji}^{(1)} x_{i,t} \right) \right). \tag{2}$$

In this case the nonlinearity is a *tanh* function, but a logistic function or any similar compressive nonlinearity could be used, depending on the requirements of the problem. The output layer neurons $y_n$ make a linear sum of the weighted inputs from the hidden layer. If an event output is required, the output neurons' continuous linear outputs can be processed by thresholding to produce event streams $z$:

$$z_n(t) = Boolean(y_n(t) > \theta) \tag{3}$$

where $\theta$ is a threshold chosen to maximize the margin of classification between event and non-event.

## 5  Solving the Output Weights

In a conventional ELM the weights $w_{nj}^{(2)}$ that connect the hidden and output layers are computed using singular value decomposition (SVD), in order to solve the linear regression between hidden layer outputs and the output layer outputs.

In a recent work, the authors have developed an online method for rapid computation of this solution in large data sets [13]. We have used this method in all the problems presented here, but note that it produces the same pseudoinverse solution as singular value decomposition, but with more moderate requirements in terms of memory and computational power.

In either case, the solution is by a pseudoinverse method, as follows. We can represent the sampled outputs of the hidden layer neurons as a time series $A = [a_1 \cdots a_k]$ where $A \in \mathbb{R}^{M \times k}$, where the columns $a$ are the outputs of the hidden layer at various times in the series. There will be an equivalent time series $Y = [y_1 \cdots y_k, Y \in \mathbb{R}^{N \times k}$ consisting of the linear outputs $y$ of the output layer corresponding to each sampled $a$. The solution required is to find the set of weights $W \in \mathbb{R}^{N \times M}$ that will minimize the error in:

$$WA = Y. \tag{4}$$

Given that the output neurons are linear, this is a linear regression problem and may be solved analytically by taking the Moore-Penrose pseudoinverse $A^+ \in \mathbb{R}^{k \times M}$ of $A$:

$$W = YA^+. \tag{5}$$

As stated, if *A* and *Y* are data available in batch form, then the pseudoinverse can be calculated by means of SVD or QR decomposition. In cases where the data sets are too large for convenient computation, or the system is online and the solution must be computed anew each time a new input-output vector pair becomes available, then an online pseudoinverse solution must be used – two such methods are the Online PseudoInverse Update Method (OPIUM) [13] and the Online Sequential ELM (OS-ELM) [20]. Given that this is intrinsically a convex optimization problem, there are many algorithms which could produce a useful solution

In processing event-based signals, we will require a further step if an output in the form of an event is required. We need to produce a binary output (event or no event) in time. This can be viewed and treated as a classification output. A simple way of doing this is to model the target output signals as time series of binary values, say *0* for no event and *1* for an event. The network weights are computed to produce these binary outputs. In use, the outputs of the linear output neurons are then fed to comparators with an appropriate threshold, in order to produce actual binary valued outputs.

In practice, a problem can arise in that if the output events are sparse in time – which they typically are – then there is very little energy in the target signal; for example, if the occurrence of events is on average one per thousand time steps, then a regression algorithm that concludes that no events ever occur will be right 99.9% of the time. This can be overcome by increasing the energy of the target events, either by increasing their nominal amplitude or length in time (expanding the length appears to work better, but we know of no mathematical or heuristic foundation for this observation). In ELM, the number of hidden layer elements should be kept smaller than the number of training data sets in order to avoid overtraining; in SKIM one would expect a similar effect to occur but given the added complexity of the data (it has a time dimension with a range that may extend to positive infinity in the case of IIR synaptic filters) it is not feasible to articulate a simple rule for the threshold hidden layer size.

## 6 Mixed-Domain Processing: Events in Combination with Continuous Valued Signals

There are modeling problems in which it may be required to include both events and continuous-valued signals. For example, neural networks in the brain, which signal by means of spike events, are also affected by the relative concentrations of neuromodulatory chemicals such as glutamate, which can be represented as continuous-valued signals.

One of the great advantages of the ELM and SKIM approach is that it allows considerable latitude in the input and hidden layers (expressed well in the title of Rahimi and Recht's report, "Weighted sums of random kitchen sinks" [14]). In the case of SKIM, we can add continuous-valued signals in the synaptic kernels, by summing them after the filter function but before the compressive nonlinearity (see figure 2). This parallel path acts exactly like a conventional ELM, and is not biologically unrealistic in the case of SKIM models that use continuous-valued signals to represent ionic or neurotransmitter concentrations in mammalian neural networks.

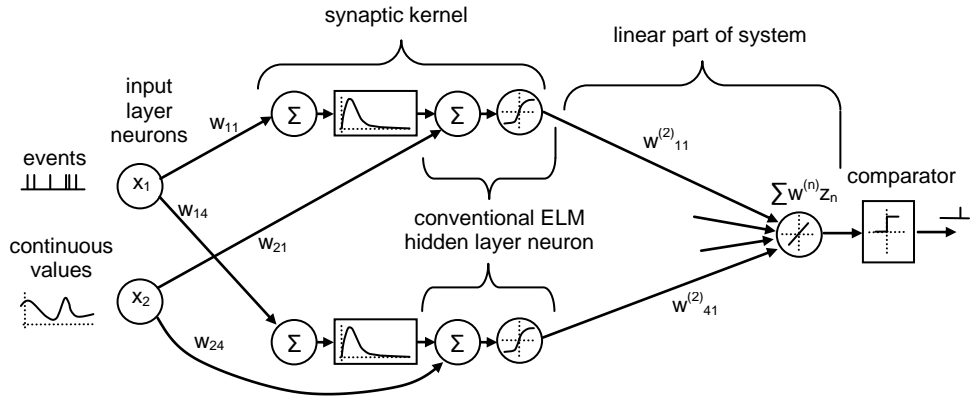

Figure 2. Event-based paths can be combined with continuously-valued signals to implement mixed-domain processing, in which the continuous-valued path is a conventional ELM in parallel with the synaptic kernel path.

## 7   Recurrent Connections and Memory

The conventional ELM network does not require explicit memory, as it is generally used to process ordered data sets rather than events in time. Networks for processing spatio-temporal patterns do require memory, as there is a necessity to remember the times and places of earlier events, in order to make decisions during the processing of later events. Most of the prior methods discussed in Section 2 make use of recurrent (feedback) connections to implement memory. This approach has some advantages. Most importantly, it eliminates the need for computational memory; the state of the network at any given moment encodes the effect of past history on future processing. It also allows for a multiplication of nonlinear effects, in that signals cycle through nonlinear elements many times (potentially an infinite number of times, with infinite-impulse response filters), and this creates rich system dynamics and complex behaviour. Finally, the intuition or vision provided by a pool of signals, resonating or echoing for some time after excitation, is very evocative for workers focusing on cognitive systems.

We have chosen to avoid the use of recurrent connections for the sake of simplicity, and from a sense that the advantages listed above may not make much contribution to real performance. Networks of recurrently-connected neurons are hampered by two overlapping requirements – they should not self-excite to the point of instability, and they should meet the fading memory requirement, which rules out a great deal of chaotic behavior. (Chaotic recurrently-connected networks are of course widely studied from a dynamical systems perspective, but not for the purposes of spatio-temporal pattern recognition.) These two requirements force practitioners to limit the recurrent gains in a network to a regime in which the dominant dynamics are exponential decay of excitation amplitudes, with some damped second order (resonant) behavior.

While we do not dispute that there is still potential for rich and complex behavior in stable recurrent networks with fading memory, if we can achieve useful memory and nonlinearity without recurrence, it significantly reduces the computational burden in practical network use. For example, ELM practitioners will be familiar with the practice of computing the weighted sums of input signals (at the hidden layer) for a complete data set, by means of a single matrix multiplication; this is manifestly not possible with a recurrently connected nonlinear hidden layer. It could be argued that the feedforward synaptic functions in SKIM require explicit memory of event times, but we have found that this can be avoided by substituting a truncated version of the synaptic kernel for each event as it is processed.

A further property of all-to-all recurrent connections is that they permit lateral inhibition in a layer, which is the foundation of winner-take-all networks and has significant computational power [21]. We consider this property to perhaps be a significant loss in our feedforward model of SKIM, and are currently evaluating its utility in deep SKIM-type networks.

## 8   An Example of the Method

There are very few benchmark problems for spatio-temporal pattern recognition; the nearest that we have found is the *mus silicium* problem posed by Hopfield and Brody [16, 17]. We have used an early variant of the SKIM method to tackle this problem, and the successful result is reported elsewhere [22]. For the purposes of illustrating the method here, we will use a generic problem that has features of many different types of application.

The general framework of this problem is that of attentional switching in a neural circuit. In recent years it has been shown in both animals and humans that areas of auditory cortex display plasticity in response to changes in attention; that is to say, an area which is showing one kind of receptivity in a passive attentional state will show a different sensitivity in an active attention state [23]; or, the pattern of sensitivity in attending to one stimulus stream will switch when attending to another stimulus stream [24]. It is hypothesized that the switching may be caused by changes in concentration of neurotransmitters such as NMDA [25].

In this model we will use a continuous-valued signal (the neurotransmitter) to switch the attention of a SKIM network from detecting one spatio-temporal pattern of events (nominally, the neural signature of a "word"), to recognizing a different spatio-temporal pattern (a different "word"). The example thus illustrates both simple event-based processing and mixed-domain processing. In one attention state the network must recognize one word, and not the other; in the other attention state, it must do vice-versa.

For the sake of visible simplicity we will use just 6 input channels, representing 5 event channels and a continuous neurotransmitter channel. The input events will consist of a mixture of two target patterns (two "words") together with some noise events generated according to a Poisson distribution.

There are two output event channels, each of which represents the occurrence of a target pattern (word) in the input streams, with switching of attention from one pattern to the other triggered by changes in the level of the neurotransmitter signal.

The network has the structure shown in Figure 2, with 5 event input neurons and one continuous input neuron, feeding 250 hidden layer synaptic kernel/neuron combinations, and being linked by trained or solved weights to two output event neurons. The synaptic kernels used were Gaussian responses with explicit delays. Using the OPIUM method for output weight solution [13], it was possible to simulate this network in 32-bit MATLAB on a notebook PC, for sequences of up to $10^5$ timesteps; simulations took less than a minute.

### 8.1 Spatio-Temporal Receptive Fields

If we train a network to recognize a particular pattern, and it emits events in response to that pattern and not others, then we can assume that the network is receptive to that pattern. However, we cannot always be sure that it is in fact the input pattern to which the network is receptive; most researchers are well familiar with systems which appear to be responding to a target signal but which are in fact responding to some artifact which has some co-occurrence with the target. We can test what it is that a network is in fact responding to, by inputting random noise signals, recording the system output events, and then reverse-correlating the output with the input to determine what distribution of input signals caused the system to output an event [26]. This input distribution is called the spatio-temporal receptive field and it is a powerful tool for analyzing "black box" systems where only the input and output are known; it is a kind of multi-dimensional transfer function.

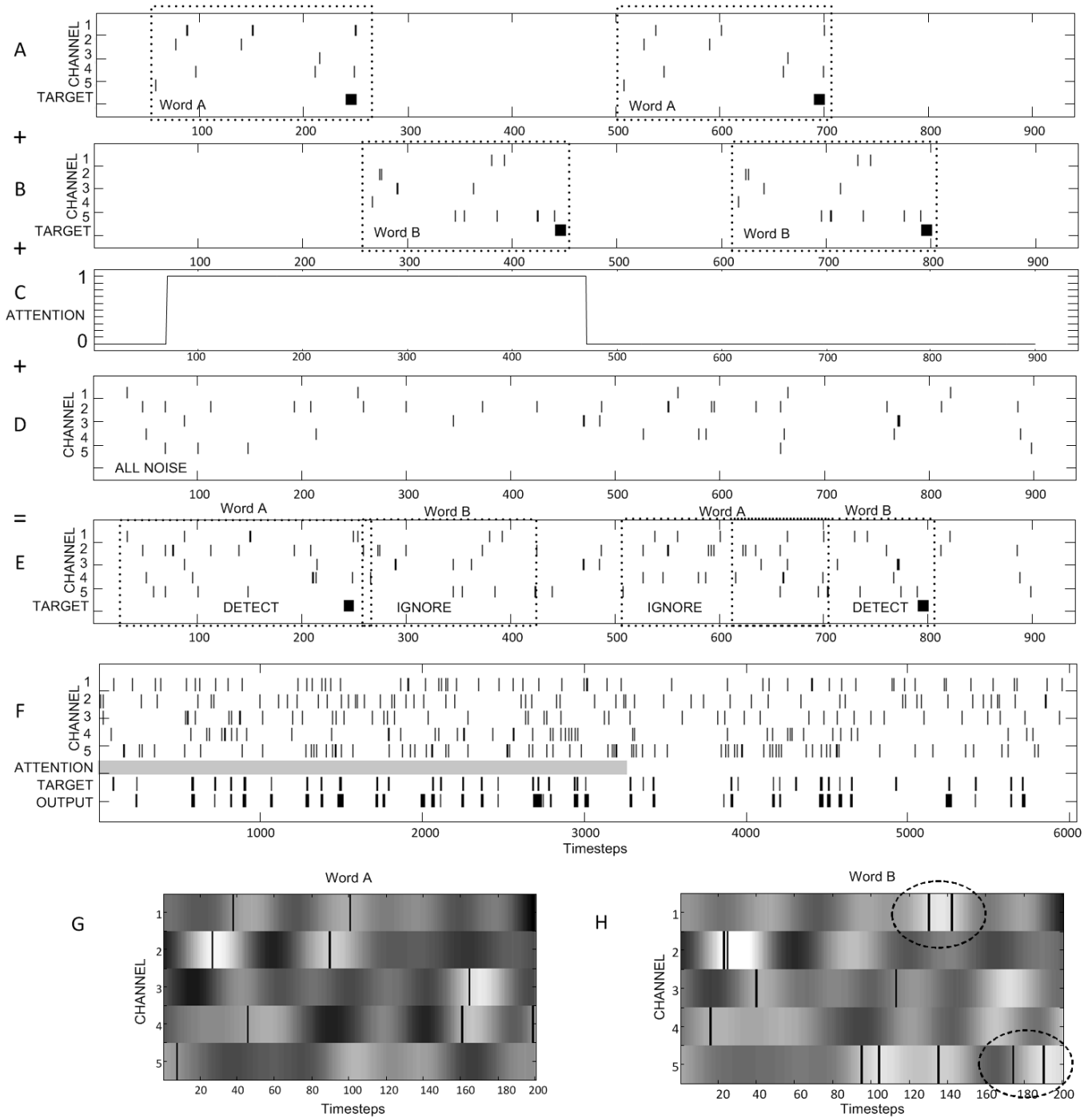

Figure 3. Typical signals in the mixed-domain network described in Section 8. Panels A through E show how a training data set is constructed, and F shows testing output. The horizontal axis indicates timesteps in all cases. A and B show raster patterns of events for two words; two occurrences of word A occur in panel A and two occurrences of word B occur in panel B. The sixth channel in each of these rasters is the target value, i.e. the output event that would be generated if that word is detected. The targets are non-unitary width (wider that the input events) in order to increase the energy in the target signal. C shows the continuous-valued attention signal, which switches the network detection attention from word A to word B. D shows Poisson-distributed noise events which are added to make the detection problem non-trivial. E shows the combined signals with a single target channel, representing the correct output for the given state of attention; with positive attention

values, word A must be detected, and with negative attention values word B must be detected. Panel F shows typical event trains during testing (note the change of timescale form previous panels). The network output is the lowest channel in F. The target event channel is immediately above it. Attention is shown in grey scale and is positive when grey, negative when white. It can be seen that the output events closely match the target events, managing the switch of attention between the two candidate words. Panels G and H show the spatio-temporal receptive fields (STRFs) for the two attentional states, with positive receptivity shown in white and negative in black. The word raster patterns are overlaid on the STRFs to indicate the positions of the word events. It can be seen that the STRFs are quite similar, as many target events are coincidentally closely placed in both words, but in two key areas (circled in word B) the network has learnt to discriminate between the two words and detect a number of events which are not common to both words. This variation in the STRFs is driven by the attentional signal.

In Figure 3, the STRFs for the two attentional states are shown. It can be seen that the network has learned both the common features of the two words, in order to distinguish them from background noise, and also some differentiating features, in order to be able to detect one but not the other depending on the attentional state. The match between the STRFs and the patterns gives us confidence that it is in fact the patterns that the network has learnt to detect, and not some artifact (such as the periodicity or average of signal power in the combined channels).

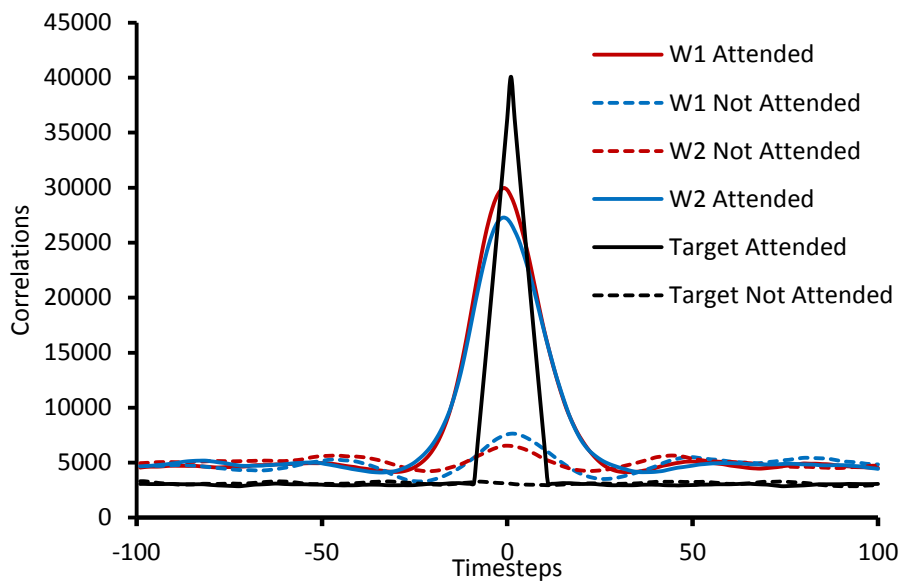

Figure 4. Cross-correlations of target and output signals, for the four permutations of attention and words, and two reference (ideal) cases. Solid lines indicate correct outputs, and dotted lines indicate error outputs. The black lines are references, representing correlation of the target (ground truth) with itself, i.e., a perfect response – note that target events were 10 timesteps in width. It can be seen that the network response was very accurate, although the small peaks in dotted lines at $t=0$ show that some words were recognized in the incorrect attention state, and the raised baselines indicate some out-of-phase detections. The data consisted of ten trials with different words in each trial, for $10^5$ timesteps each; there were a total of 34206 words processed. Poisson noise events at a mean rate of 0.01/timestep were added to each channel during these trials, so that approximately 50% of all events were noise.

It is clear from panel F in Figure 3 that the network is capable of recognizing the target patterns, and switching attention from one pattern to another on short timescales without being retrained. Figure 4 gives a measure of the accuracy of the model. We have correlated the target (ground truth) signal with the output signal, for the four possible outcomes (word A recognized during attention on word A; word A recognized during attention on word B – a wrong outcome; and so on). The correlations indicate that the network is recognizing the words, in the appropriate attention state, with a high degree of accuracy.

## 9 Conclusions

The following features of the SKIM approach illustrate its utility in systems modeling and machine learning:

- The method does not make use of event-firing neurons in the hidden layer, but achieves nonlinear separation through the use of synaptic kernels. While these have much of the same intrinsic function as ELM hidden layer neurons, the structure is significantly different to reservoir methods such as the liquid state machine, where all neuronal elements are event-producing (spiking) neurons. In most systems, the emission of an event requires or consumes energy, so a network in which events are not emitted for the purposes of internal representation of states is likely to be more economical than one which uses spikes for this purpose.
- The method uses feedforward-only paths, and achieves memory by means of synaptic kernel filter responses, rather than by use of recurrent connections. This avoids the problems of stability, separability and fading memory that must be addressed in recurrent spiking networks.
- The method is able to combine event-based and continuous signals in mixed-signal domain processing. This significantly improves its utility in modeling many real-world problems, and particularly the action of neurotransmitters in biological spiking networks.
- The use of the ELM structure allows enormous flexibility in the structuring of the hidden layer; this structure is essentially independent from the training or solution part of the network (the output layer), although the quality of the output is obviously affected by it. It allows practitioners to choose a hidden layer structure that most closely maps to the problem space, without being concerned that training will become impossible.

In comparison with existing models, we note the following features of SKIM that differentiate it, and which may help practitioners to choose an appropriate method:

- Compared to the general class of reservoir models such as the Liquid State Machine and Echo State Network, which are characterized by a hidden layer with recurrent connections, SKIM is required to have explicit memory and cannot be characterized as a state machine. However, avoiding the use of recurrent connections gives significant advantages in terms of simplicity and stability. In addition, SKIM offers the use of different kernel types which may be more appropriate or realistic when biological neurons (or other physically real systems) are being simulated.

- The Neural Engineering Framework is a system for synthesizing large cognitive meta-networks, and hence cannot be directly compared to SKIM. Its core algorithm uses ensembles of spiking neurons which are interconnected by means of pseudoinversely-solved weights. In principle, the NEF could use SKIM networks for its sub-modules, and this ought to offer significant economies in network complexity as it would replace spiking neurons with synapses.
- Tempotron networks are very similar to a particular case of SKIM in which synapses are represented as a sum of exponentials, although the use of an incremental learning rule in the Tempotron is a significant departure. The pseudoinverse solution in SKIM allows very rapid synthesis of networks, with a fast and reliable solution for the weights, and the extension to multiple types of synapse and the provision of continuous signals in mixed-domain situations renders the SKIM approach more versatile.
- Deltron networks use learning of connection delays rather than weights to achieve spatio-temporal pattern recognition. This is a more direct method of tackling the problem. It has two issues which may be more or less problematic to practitioners, depending on their purposes: it uses a gradient learning rule to find the correct delays, and current paradigms of mammalian neural networks suggest that delay adaptation occurs by means of synaptic weight changes. Nonetheless it is an interesting method that may have many applications.

In practice, users will select the network that most suits their application. SKIM offers rapid synthesis, stability, accuracy, and a significant versatility in synaptic representation. We consider that the SKIM model will be very useful in computational neuroscience, particularly as models move from single-compartment simplicity to higher levels of detail.

**References**

[1] G.-B. Huang, Q.-Y. Zhu, and C.-K. Siew, "Extreme learning machine: Theory and applications," *Neurocomputing*, vol. 70, pp. 489–501, 2006.

[2] K.J. Astrom and B.M. Bernhardsson, "Comparison of Riemann and Lebesgue sampling for first order stochastic systems," *Decision and Control, Proceedings of the 41st IEEE Conference on*, vol. 2, pp. 2011 – 2016, 2002.

[3] P. Dayan and L.F. Abbott, *Theoretical Neuroscience: Computational and Mathematical Modeling of Neural Systems*, MIT Press, Boston, 2001.

[4] H. C. Tuckwell, *Stochastic Processes in the Neurosciences*, SIAM, New York, 1989.

[5] R. Van Rullen, and S.J. Thorpe, "Rate Coding Versus Temporal Order Coding: What the Retinal Ganglion Cells Tell the Visual Cortex," *Neural Computation*, 13:6, 1255-1283, 2001.

[6] E.R. Kandel and J.H. Schwartz, *Principles of Neural Science,* $2^{nd}$ Edition, Elsevier, New York, 1985.

[7] W. Maass and H. Markram, "On the Computational Power of Recurrent Circuits of Spiking Neurons", *Journal of Computer and System Science,* vol. 69, pp. 593–616, 2004.

[8] H. Jaeger, "Adaptive nonlinear system identification with echo state networks", in *Advances in Neural Information Processing Systems 15*, S. Becker, S. Thrun, K. Obermayer (Eds), (MIT Press, Cambridge, MA, pp. 593-600, 2003.

[9] W. Maass and E.D. Sontag. "Neural Systems as Nonlinear Filters," *Neural Comput.*, vol. 12, pp. 1743-1772, 2000.

[10] C. Eliasmith and C. H. Anderson, *Neural Engineering: Computation, representation and dynamics in neurobiological systems*. MIT Press, 2003.

[11] C. Eliasmith, T.C. Stewart, X. Choo, T. Bekolay, T. DeWolf, Y. Tang and D. Rasmussen, "A Large-Scale Model of the Functioning Brain," *Science*, vol. 338, pp. 1202-1205, 2012.

[12] E.M. Izhikevich, *Dynamical Systems in Neuroscience: The Geometry of Excitability and Bursting*, MIT Press, Cambridge, 2010.

[13] J. Tapson and A. van Schaik, "Learning the pseudoinverse solution to network weights," *Neural Networks*, vol. 45, pp. 94-100, 2013.

[14] A. Rahimi, and B. Recht, "Weighted sums of random kitchen sinks: Replacing minimization with randomization in learning," In D. Koller, D. Schuurmans, Y. Bengio, & L. Bottou (Eds.), *Advances in neural information processing systems, 21,* pp. 1313–1320, MIT Press, Cambridge, 2009.

[15] J.J. Hopfield and C.D. Brody, "What is a moment? "Cortical" sensory integration over a brief interval," *Proceedings of the National Academy of Sciences (PNAS)*, vol. 97, pp. 13919-13924, 2000.

[16] J.J. Hopfield and C.D. Brody, "What is a moment? Transient synchrony as a collective mechanism for spatiotemporal integration," *Proceedings of the National Academy of Sciences (PNAS)*, vol. 98, pp. 1282–1287, 2001.

[17] R. Gütig and H. Sompolinksy, "The tempotron: a neuron that learns spike timing-based decisions", *Nat. Neurosci,* vol. 9, pp. 420-428, 2006.

[18] S. Hussein, A. Basu, M. Wang and T.J. Hamilton, "DELTRON: Neuromorphic architectures for delay based learning", *Circuits and Systems (APCCAS), 2012 IEEE Asia Pacific Conference on*, pp. 304 – 307, 2012.

[19] S. Hussein, A. Basu, R. Wang and T.J. Hamilton, "Delay Learning Architectures for Memory and Classification", *Neuroengineering*, in press, 2013.

[20] N.-Y. Liang, G.-B. Huang, P. Saratchandran, and N. Sundararajan, "A fast and accurate online sequential learning algorithm for feedforward networks.," *IEEE Trans. Neural Networks*, vol. 17, no. 6, pp. 1411–23, Nov. 2006.

[21] W. Maass, "On the Computational Power of Winner-Take-All," *Neural Computation*, vol. 12, pp. 2519-2535, 2001.

[22] J. Tapson, G. Cohen, S. Afshar, K. Stiefel, Y. Buskila, R. Wang, T.J. Hamilton and A. van Schaik, "Synthesis of neural networks for spatio-temporal spike pattern recognition and processing," *Frontiers in Neuroscience*, vol. 7:153, doi:10.3389/fnins.2013.00153, 2013.

[23] J. Fritz, S. Shamma, M. Elhilali, and D. Klein "Rapid task-related plasticity of spectrotemporal receptive fields in primary auditory cortex," *Nat. Neurosci*, vol. 6, pp. 1216-1223, 2003.

[24] N. Mesgarani and E.F. Chang, "Selective cortical representation of attended speaker in multi-talker speech perception," *Nature*, vol. 485, pp. 233-236, 2012.

[25] J. Fritz, M. Elhilali, S. David and S. Shamma "Auditory attention-focusing the searchlight on sound," *Current Opinion in Neurobiology*, vol. 17, pp. 437-455, 2007.

[26] F. Rieke, D. Warland, R. De Ruyter van Stevenink, and W. Bialek, *Spikes: Exploring the Neural Code*, Bradford, Cambridge, 1999.